%% file: anant.tex
\documentclass[12pt]{article}

\input preamble.tex

\title{Social Media Bot Detection using Dropout-GAN}

\author{Anant Shukla\footnotemark[1]\ \ \ 
Martin  Jure\vvv{c}ek\footnotemark[2]\ \ \ 
Mark Stamp\footnotemark[1]\,\,\footnotemark[3]}

\begin{document}

\symbolfootnotetext[1]{Department of Computer Science, San Jose State University}
\symbolfootnotetext[2]{Faculty of Information Technology, Czech Technical University in Prague}
\symbolfootnotetext[3]{mark.stamp$@$sjsu.edu}

\maketitle

\abstract
Bot activity on social media platforms is a pervasive problem, undermining the credibility of online discourse 
and potentially leading to cybercrime. We propose an approach to bot detection using 
Generative Adversarial Networks (GAN). We discuss how we overcome the issue of mode collapse by utilizing multiple discriminators to train against one generator, while decoupling the discriminator to perform social media bot detection and utilizing the generator for data augmentation. In terms of classification accuracy, our approach outperforms the state-of-the-art techniques in this field. We also show how the generator in the GAN can be used to evade such a classification technique.

\section{Introduction}

Social media platforms have transformed the way we communicate, interact, and consume information. 
However, the issue of bot activity on these platforms has become a pervasive problem, undermining 
the credibility of online discourse and potentially leading to cybercrime. Here, ``bots'' refer to computer 
algorithms that mimic human users, producing content, such as posting, liking, 
or retweeting content, and interacting with users~\cite{ferrara2016rise}. 

Twitter,\footnote{Twitter has been recently renamed $\mathbb{X}$. We refer to it as Twitter throughout this paper.}
a prominent microblogging platform, has garnered significant attention in recent years due to its pervasive problem with social media bots. This phenomenon, deeply rooted in the landscape of online communication, has become a focal point of academic research and technological discourse~\cite{cresci2023demystifying}. The proliferation of bots on Twitter has raised concerns about the integrity of information dissemination, the authenticity of user interactions, and the potential for algorithmic manipulation of public discourse. 

Generative Adversarial Networks (GAN)~\cite{goodfellow2014generative} have recently gained popularity in the field of machine learning due to their ability to generate realistic synthetic data. GANs have also shown promise for their classification accuracy when attempting to distinguish users as humans or bots in a social media setting~\cite{najari2022ganbot}. 

In this paper, we propose and analyze 
an approach to bot detection on social media platforms using GANs. Our method involves training a GAN 
on real and synthetic data to distinguish between genuine users and bots. We demonstrate that our approach outperforms---in terms of accuracy---the state-of-the-art techniques in this field. 
Furthermore, we demonstrate how this approach can be confounded by training another generator in a ``Dropout-GAN'' architecture, which will consistently generate bot behavior that the discriminator cannot accurately classify. 

Our primary contributions can be summarized as follows: 
\begin{itemize}
\item Accurate bot detection --- We demonstrate that GANs can be trained to accurately detect bots on social media. 
Specifically, we highlight how the GAN discriminator can be used for such classification. 
\item Novel optimization for bot detection --- We show how GANs can be further improved by using a framework that strengthens the GAN generator, to help train better performing GAN discriminators for bot detection.
\item Improved bot behavior prediction --- We show that the GAN generator can be used to defeat our classification model by training against multiple discriminators to prevent the generator from overpowering the discrimnator.  
This can be viewed as a possible preview of how social media bots may behave in the future, and it
points to directions for further research.
\end{itemize}

The remainder of this paper is organized as follows. In Section~\ref{chap:background}, 
we provide a brief overview of related work in bot detection. Section~\ref{chap:impl} 
describes our proposed approach using GANs for bot detection, and we discuss our experimental setup. 
Section~\ref{chap:res} presents our results. Finally, in Section~\ref{chap:conc}, 
we conclude with a summary of our findings and we consider future research directions.

\section{Background}\label{chap:background}

In this section, we first provide an overview of relevant related work. Then we introduce the various learning
models that are considered in this paper, with an emphasis on GANs.

\subsection{Related Work}

Various techniques have been proposed for detecting bots on social media platforms, including user 
behavior analysis, content analysis, and network analysis~\cite{orabi2020detection}. 
One of the well-known works in user behavior analysis is the BotOrNot system~\cite{davis2016botornot} which relies on over a thousand features to determine whether a Twitter user's behavior resembles that of a bot. This work was later expanded upon in~\cite{varol2017online} where the authors trained the model on a new dataset, and demonstrated which features they leveraged in their technique. 

The Random Forest (RF) technique saw early success, as detailed in~\cite{david2016features}. Another technique previously used for bot detection is boosting. For example, in~\cite{morstatter2016new} a modified version of AdaBoost called BoostOR was used with success, where BoostOR is optimized for~F1 score by balancing the precision and recall of the classifier. Convolutional Neural Networks (CNN) have also been applied, with one example of such research 
being~\cite{cai2017detecting}. 

A clustering-based approach is considered in~\cite{gera2022t} where a Centroid Initialization Algorithm is used to identify malicious bot users. 
Some previous work has explored graph-based techniques~\cite{zhao2020multi}, while others use deep learning models such as Long Short-Term Memory (LSTM)~\cite{kudugunta2018deep} in an attempt to deal with the challenges associated with bot detection. In addition, ensemble techniques have also been proposed, in tandem with appropriate feature selection~\cite{shukla2021enhanced}. Other research efforts have investigated the utilization of Natural Language Processing (NLP) techniques, including word embeddings~\cite{pennington2014glove}, for improved feature encoding and representation. These prior investigations have shed light on the effectiveness of different methodologies and paved the way for further advancements in the realm of bot detection and analysis. Building upon these existing foundations, this study aims to contribute novel insights by employing an extensible open framework that integrates multiple discriminators in the training process.

Work has been previously done using GANs to identify social media bots~\cite{najari2022ganbot}. 
However, in our approach, we propose a simpler architecture, and we expand the concept of using GANs 
for social media bot detection to a multilingual dataset. The dataset we use includes tweets from~54 languages, 
and our method also outperforms the previous work in~\cite{najari2022ganbot}, in terms of accuracy and other relevant metrics such as precision, recall, F1 score, and so on.

\subsection{Bot Detection Techniques}

Bot detection techniques can be broadly classified into two categories: 
feature-based techniques and graph-based techniques. 
Feature-based techniques rely on features such as user behavior, content, and so on,  
whereas graph-based 
techniques emphasize the use of graph analysis to detect bot activity.
Such graph analysis may focus on aspects such as network properties, graph structure, and spatial information like proximity to other humans or bots~\cite{orabi2020detection}.

\subsubsection{Feature Based Techniques}

Feature-based techniques for bot detection on social media platforms rely on features such as user behavior, 
content, and network properties. AdaBoost, Random Forest, 
Decision Trees, Support Vector Machines (SVM), 
$k$-Nearest Neighbors ($k$-NN), Multilayer Perceptrons (MLP), 
and Logistic Regression \cite{orabi2020detection} are examples of feature-based techniques that have been used in bot detection. 
These techniques are trained on a set of labeled data, where the features are extracted from user behavior, 
content, and other properties such as tweet metadata. The trained classifier then uses these features to distinguish 
between genuine users and bots. 

Feature-based techniques are used for bot detection due to their ability to handle high-dimensional data and their 
ability to detect complex relationships between features. However, these methods are susceptible to sophisticated 
bots that may modify the features that are scrutinized for detection, and thereby evade feature-based 
detection methods~\cite{cresci2017paradigm}.

\subsubsection{Graph Based Techniques}

Graph-based techniques for bot detection on social media platforms use graph analysis to detect bot activity. 
Such techniques often work by analyzing the structure of the social media network and identifying patterns 
of bot activity, such as bots that are part of a botnet or bots that are using the same~IP address. These techniques 
can detect sophisticated bots that are difficult to detect using feature-based techniques. However, graph-based 
techniques are generally computationally expensive, and may not be feasible for larger graphs~\cite{zhang2020deep}.
The following graph-based learning techniques have been applied to the 
bot detection problem:
\begin{itemize}
\item Graph Convolutional Networks (GCN)
\item GraphSAGE (SAmple and aggreGatE)
\item Graph Attention Networks (GAT)
\item Heterogeneous Graph Transformers (HGT) \cite{hu2020HGT}
\item Hamiltonian Generative Networks (HGN) \cite{toth2019hamiltonian}
\item BotRGCN \cite{feng2021botrgcn}
\item Relational Graph Transformers (RGT) \cite{feng2022heterogeneity}
\end{itemize}

\subsection{GAN}

Generative Adversarial Networks (GANs) are a class of deep learning models that were introduced 
in~\cite{goodfellow2014generative}. GANs consist of two components, namely, a generator and a discriminator. 
The generator produces synthetic data that is meant to resemble real training data, while the discriminator 
tries to distinguish between real data and the synthetic data produced by the generator. 

GANs have been applied to a wide range of tasks, such as image and text generation~\cite{kaddoura2023primer}, and anomaly detection~\cite{schlegl2019f}. 
In the context of bot detection on social media platforms, GANs can be used to model the behavior of bots by 
training the generator to produce synthetic bot behavior that resembles real bot behavior.

The discriminator in the GAN can be used to detect social media bots by training it to distinguish between genuine 
user behavior and synthetic bot behavior. The discriminator can be trained on a dataset of labeled data where the 
labels indicate whether a user is a bot or not. The discriminator then learns to identify patterns in the data that are 
characteristic of bot behavior. Once the discriminator is trained, it can be used to detect bots in new data that it 
has not seen before.

The generator in the GAN can be used to model social media bots by training it to produce synthetic bot behavior that resembles real bot behavior. The generator is trained on a dataset of labeled data where the labels indicate whether a user is a bot or not. The generator then learns to produce synthetic bot behavior that is similar to the behavior of real bots in the dataset. This synthetic data can serve to strengthen the discriminator, while the
generator-produced data provides additional examples of potential bot behavior.

\section{Implementation}\label{chap:impl}

In this section, we first discuss the dataset used in our experiments. We then turn our attention 
to the training of the GAN models that are the focus of our research.

\subsection{Dataset}

The MGTAB~\cite{shi2023mgtab} dataset exhibits substantial potential as a foundation for bot research due to its remarkable wealth of data. The authors created a new graph-based dataset by selecting~100 seed accounts actively engaged in discourse about Japan's plan to release nuclear wastewater into the ocean, then collecting the~10,000 most recent tweets for each user, eventually accumulating a total~1,554,000 users and~135,450,000 tweets. After cleaning, they retain~410,199 users and roughly~40 million tweets. However, not all these users and tweets are labeled; only~10,199 users and their associated tweets were manually annotated to describe whether they were bot accounts or human accounts, and also have other labels that are relevant for stance detection (``neutral'',``against'',``support''), but were not used in our research. The authors of~\cite{shi2023mgtab} employ many contemporary machine learning techniques, as well as graph-based machine learning techniques to determine their performance on the dataset. These results serve to benchmark the performance against previous datasets as well. By virtue of its versatile combination of feature-based and graph-based characteristics, this dataset is highly conducive to effective experimentation and analysis.

Twibot-22~\cite{feng2022twibot} also stands out as a notable dataset, particularly 
due to influencing the emergence of the BotRGCN technique~\cite{feng2021botrgcn}, 
which currently ranks among the most successful 
graph-based methodologies for bot detection. Notably, this technique achieves an 
impressive accuracy 
of~87.2\% on the MGTAB dataset and~79.7\% on the Twibot-22 benchmark. The 
authors built a new dataset by starting from the \texttt{@NeurIPSConf} account, and 
then using its first~1000 followers and followees as candidates for BFS expansion. 
They then randomly adopt one of two sampling strategies: distribution diversity or 
value diversity, and randomly select metadata, to include six users from its 
neighborhood into the dataset. The dataset itself 
comprises~92,932,326 users and~170,185,937 follow relationships. 
Each user has its first~1000 tweets collected as well. Like MGTAB, 
Twibot-22 also tested a variety of learning techniques which serve as benchmarks.

We selected the MGTAB dataset for our experiments. This dataset contains a mix of 
genuine users and bots. It contains both feature and graph information, but we use the feature data exclusively, 
which serves to simplify the training process. 
There are~788 features in total, with the~20 best features, 
based on Information Gain (IG), 
listed in Table~\ref{tab:MGTAB_stats}. 

\begin{table}[!htb]
\centering
\caption{Top~20 features in MGTAB based on information gain}\label{tab:MGTAB_stats}
\adjustbox{scale=0.8}{
\begin{tabular}{l|lccc}
\midrule\midrule
Feature & Description & Type & IG \\
\midrule
\texttt{followers friends ratio} & Followers/friends & Numerical & 0.391857 \\
\texttt{listed count} & Public members list  & Numerical & 0.333101 \\
\texttt{description length} & Length of description & Numerical & 0.194765 \\
\texttt{followers count} & Number of followers & Numerical & 0.176186 \\
\texttt{URL} & URL set or not & Boolean & 0.064248 \\
\texttt{name length} & Length of name & Numerical & 0.040335 \\
\texttt{created} & Time account created & Numerical & 0.034079 \\
\texttt{friends count} & Number of friends & Numerical & 0.031598 \\
\texttt{profile} & Default or not & Boolean & 0.025997 \\
\texttt{profile image} & Default or not & Boolean & 0.025402 \\
\texttt{sidebar border color} & Default or not & Boolean & 0.023105 \\
\texttt{sidebar fill color} & Default or not & Boolean & 0.022359 \\
\texttt{geographic location} & Enabled or not & Boolean & 0.019302 \\
\texttt{statuses count} & Tweets and retweets & Numerical & 0.015544 \\
\texttt{favorites count} & Account likes & Numerical & 0.011768 \\
\texttt{verified} & Verified or not & Boolean & 0.010902 \\
\texttt{background image} & Use or not & Boolean & 0.007877 \\
\texttt{screen name length} & Self-explanatory & Numerical & 0.007641 \\
\texttt{profile background URL} & Present or not & Boolean & 0.005923 \\
\texttt{profile background color} & Default or not & Boolean & 0.005841 \\
\midrule\midrule
\end{tabular}
}
\end{table}

There are~410,199 total accounts in the MGTAB data that we use in 
our experiments, and each account has 
its~10,000 most recent tweets included. 
Table~\ref{tab:dataset_details} provides more 
details on this MGTAB data.

\begin{table}[!htb]
\centering
\caption{Accounts in MGTAB dataset}\label{tab:dataset_details}
\adjustbox{scale=0.85}{
\begin{tabular}{ccc|c}
\midrule\midrule
Humans & Bots & Unlabeled & Total \\
\midrule
7,451 & 2,748 & 400,000 & 410,199 \\
\midrule\midrule
\end{tabular}
}
\end{table}

The dataset has already undergone preprocessing, utilizing 
Language-agnostic BERT Sentence Embedding (LaBSE)~\cite{feng2020language} 
for feature encoding. LaBSE expands upon the ideas of BERT~\cite{devlin2018bert} 
to multilingual and cross-lingual sentence embeddings, which works well for our use case 
since the dataset contains~54 languages. 
The LaBSE model also performs strongly on 
those languages where LaBSE does not have any explicit
training data, likely due to language similarity and the massively multilingual nature of the model.

Additionally, minmax scaling has been applied to normalize all features 
within the range of~0 to~1. This meticulous preprocessing renders the dataset readily utilizable without 
the need for further preprocessing. We divided the dataset into training, validation, and testing sets
based on an~80-10-10 split, respectively. 

\subsection{Feature Selection}

Even though the dataset has been preprocessed, we carried out some experimentation to determine which combination of features will be optimal. We experimented using different feature vector sizes, specifically, selecting~20 features, 50 features, and~100 features, based on descending order of information gain. We found that using the first~100 features ranked by information gain is optimal for our GAN  models, while other techniques we also experimented with (listed in Table~\ref{tab:performance}) seemed to benefit the most from using the first~50 features, again, ranked by information gain. 

\begin{figure}[!htb]
\centering
\adjustbox{scale=0.6}{\includegraphics{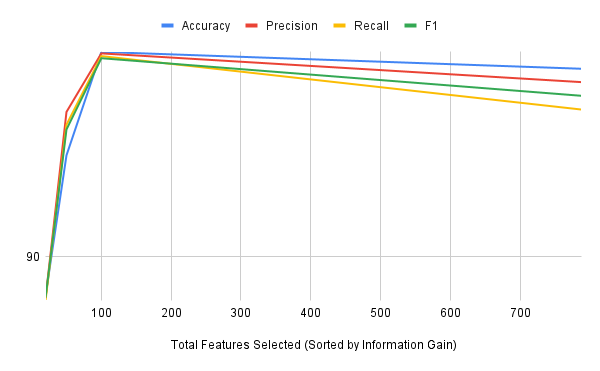}}
\caption{GAN metrics as a function of the number of features}
\label{gantrain}
\end{figure}

\subsection{Training: Regular GAN}

Figure~\ref{gantrain} depicts the initial setup that we used to train our GAN models. 
The initial goal was on training a discriminator that can distinguish between bots and humans accurately. 

Table~\ref{tab:MGTAB_Hyperparams} provides an overview of the hyperparameters explored during a 
comprehensive grid search experiment. This search aimed to determine a near-optimal 
configuration for our GAN discriminator model. The best-performing values 
(in terms of test accuracy) are presented in bold text in Table~\ref{tab:MGTAB_Hyperparams}. 
These values were obtained by performing the grid search on 
the~10,499 labeled users that MGTAB provides. 

\begin{table}[!htb]
\centering
\caption{Hyperparameters tested (selected values in boldface)}\label{tab:MGTAB_Hyperparams}
\adjustbox{scale=0.75}{
\begin{tabular}{l|l|l}
\toprule\toprule
Hyperparameter & Search space & Type \\
\midrule
\texttt{Learning Rate} & $0.001, \textbf{0.002}, \ldots , 0.01, \ldots , 0.1$ & Continuous \\
\texttt{Batch Size} & $32, 64, 128, \textbf{256}$ & Discrete \\
\texttt{Activation Function} & Sigmoid, \textbf{ReLU}, LeakyReLU & Categorical \\
\texttt{Optimizer} &  \textbf{Adam}, SGD & Categorical \\
\texttt{Noise Dimension} & 100 & Fixed \\
\texttt{Dropout Rate} & $0.0, 0.1, 0.2, \ldots , \textbf{0.5}, \ldots , 1$ & Continuous \\
\texttt{Number of Epochs} & $10, 25, \textbf{50}, 100$ & Discrete \\
\texttt{Number of Linear Layers (Discriminator)} & $1, 2, \textbf{3}, 4, \ldots , 20$ & Discrete \\
\texttt{Number of Linear Layers (Generator)} & $1, 2, \textbf{3}, 4, \ldots , 20$ & Discrete \\
\bottomrule\bottomrule
\end{tabular}
}
\end{table}

We train all of our GAN models using the PyTorch~\cite{pytorch} framework. 
From Table~\ref{tab:MGTAB_Hyperparams}, we observe that our best results
are obtained using the Adam optimizer with a learning rate of~0.002 and a batch size of~256. 
We train our models for~50 epochs.
During training, we optimized the discriminator and generator models using the 
binary cross-entropy loss function, combined with a sigmoid layer. 
The discriminator was trained to distinguish between genuine user behavior and 
bot behavior, in both real samples, and samples produced by the generator. Essentially, we use the generator as a sort of data augmentation technique. The generator was trained to produce synthetic bot behavior 
that resembles real bot behavior. We trained a discriminator that is able to predict whether a user is a human user or a bot user with 99.3\% accuracy - this was our best-performing discriminator after utilizing the optimal hyperparameters that we discovered during our grid search earlier. From now on, we will refer 
to the best-performing discriminator as D*.

\begin{figure}[!htb]
\centering
\adjustbox{scale=0.325}{\includegraphics{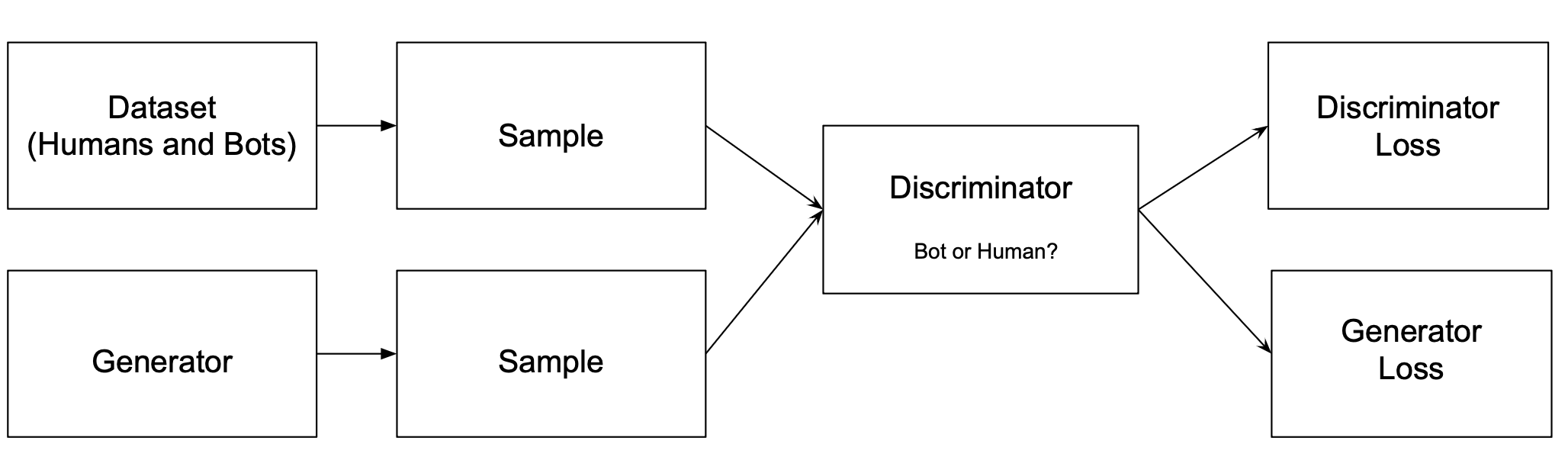}}
\caption{Training setup for GAN}
\label{gantrain}
\end{figure}

\subsection{Training: Dropout GAN}

Our investigation revealed that a conventional GAN setup employed for bot prediction demonstrates 
commendable efficacy in bot identification. However, this success stems from the discriminator 
completely overwhelming the generator---or at the very least, at a notable cost to the 
generator's performance. Around epoch~25, clear indications of mode collapse became evident, 
as the generator predominantly generated a limited subset of outputs. In our particular case, 
the generator produced synthetic bot results at a much higher rate than synthetic human results. 
Ideally, we should have about an equal chance of generating one or the other. Due to this issue of mode collapse, the discriminator D* would only guess that a potential user is a bot, which negatively impacted the~F1 score of the discriminator.

\begin{figure}[!htb]
\centering
\adjustbox{scale=0.5}{\includegraphics{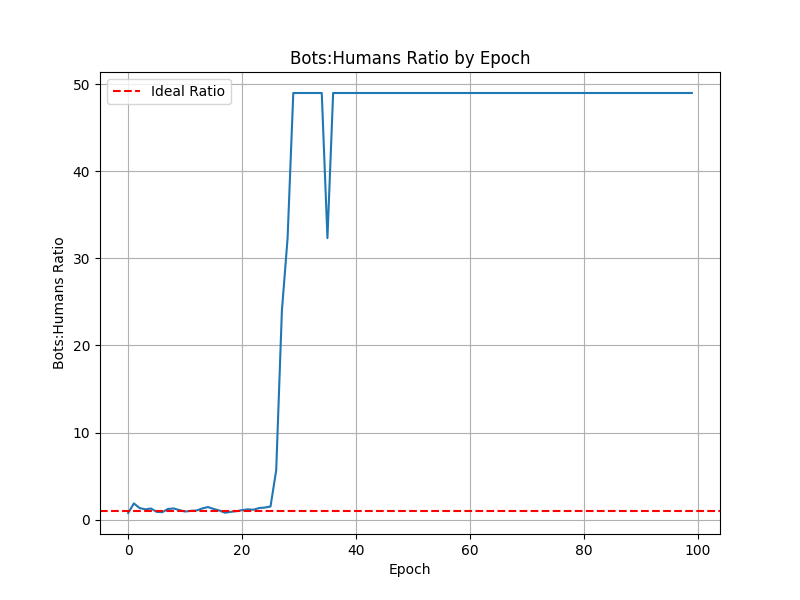}}
\caption{Mode collapse visualized}
\label{modecollapse}
\end{figure}

Figure~\ref{modecollapse} shows the bot-to-human ratio, as a function of training epoch.
Whereas the ideal ratio is about~1-to-1, from Figure~\ref{modecollapse} we observe that the
ratio is elevated after just~25 epochs, and reaches nearly~50-to-1 by epoch~30.
Consequently, the discriminator assumed that generated 
samples were bots, and no improvement in learning could then take place. 
We initially experimented with adding dropout layers to the 
discriminator and generator, but that was not effective in alleviating the problem. 
We also tried tweaking the learning rate of both the discriminator and generator, 
but that did not improve the situation either.

\begin{figure}[!htb]
\centering
\adjustbox{scale=0.225}{\includegraphics{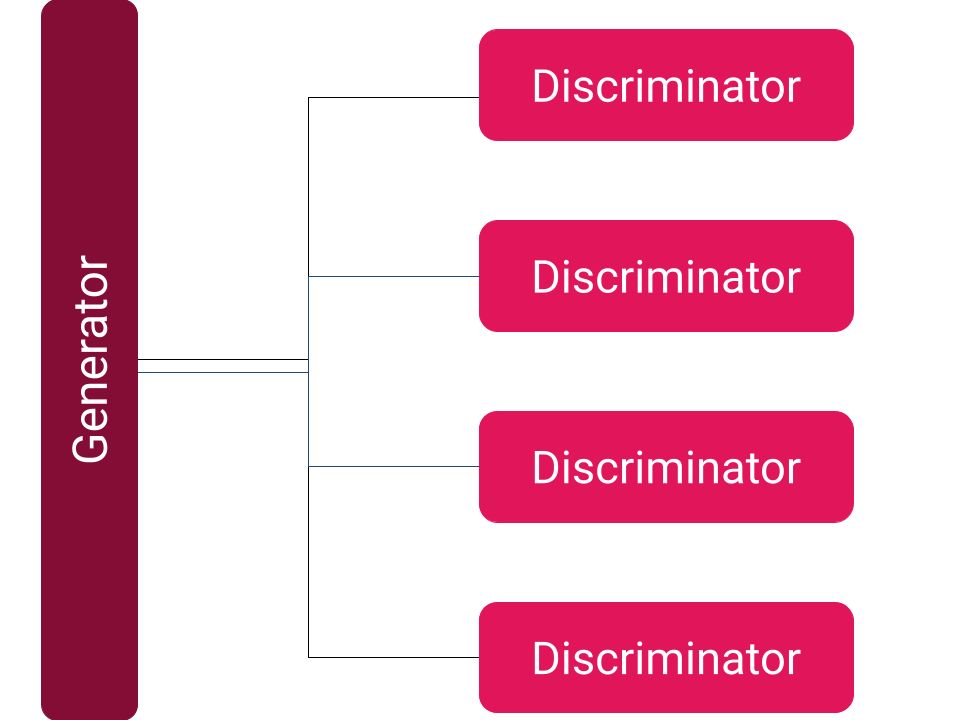}}
\caption{Dropout-GAN architecture}
\label{DropoutGAN Architecture}
\end{figure}

The research in~\cite{mordido2018dropout} inspired an approach that we found to be successful
for dealing with the problem of mode collapse. 
Specifically, we found that a ``many-discriminator, one-generator'' GAN architecture, 
which is known as a Dropout-GAN, mitigated the issue of mode collapse. 
Figure~\ref{DropoutGAN Architecture} illustrates this Dropout-GAN approach.

In contrast to conventional GAN configurations, the Dropout-GAN approach involves 
training the generator against 
multiple discriminators within each epoch. We randomly choose a subset of discriminators 
to train against the generator every epoch, which is similar to the mechanism of 
neuron deactivation, or ``dropout,'' in a conventional neural network. 
Dropout is a powerful regularization technique widely employed in 
conventional neural networks to mitigate overfitting. Dropout operates by randomly 
deactivating a fraction of neurons during each forward and backward pass of training. 
This stochastic dropout of neurons ensures that the network does not rely too heavily on 
a small subset of neurons while others atrophy. In effect, dropouts enhance generalization
and reduce overfitting by preventing the neural network from simply memorizing the training data.  

\begin{figure}[!htb]
\centering
\adjustbox{scale=0.30}{\includegraphics{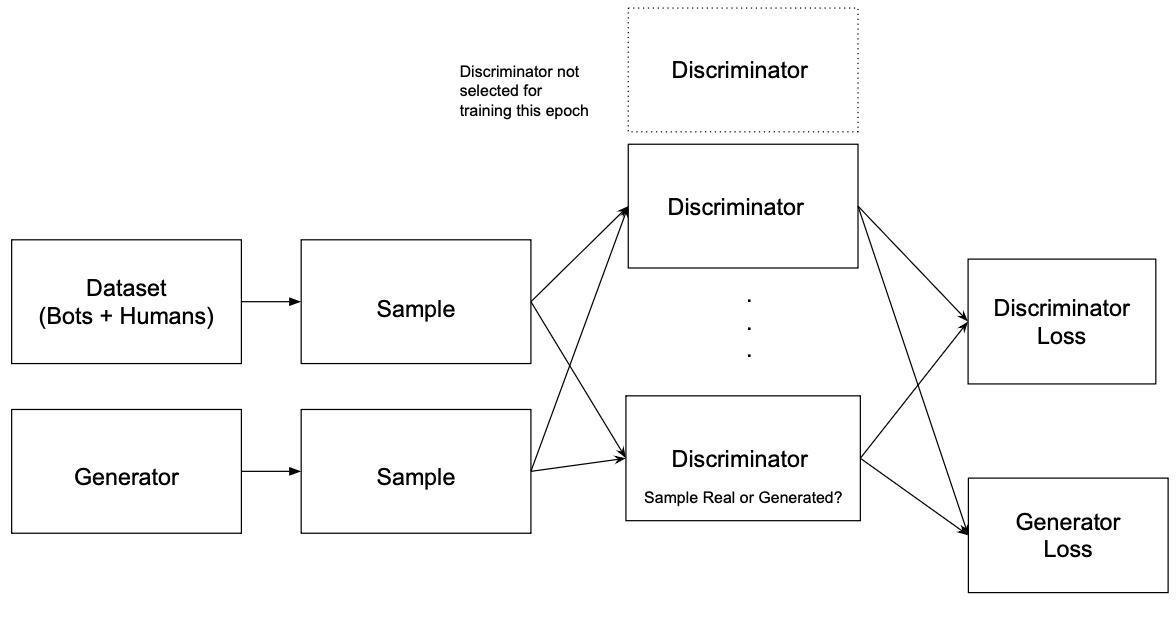}}
\caption{Training setup for Dropout-GAN}\label{fig:train_DG}
\end{figure}

In our Dropout-GAN training step, the generator is trained against a discriminator 
with a certain probability. If this probability 
surpasses a predetermined threshold, it triggers the training of the discriminator against the generator. 
Conversely, if the probability falls below the threshold, the discriminator is excluded from the training 
process, and its involvement in the training step for the epoch is effectively nullified. Per epoch,
each discriminator is randomly selected for training (or not) based on a specified threshold.
This training process is illustrated in Figure~\ref{fig:train_DG}.

\begin{figure}[!htb]
\centering
\begin{tabular}{c}
\adjustbox{scale=0.325}{\includegraphics{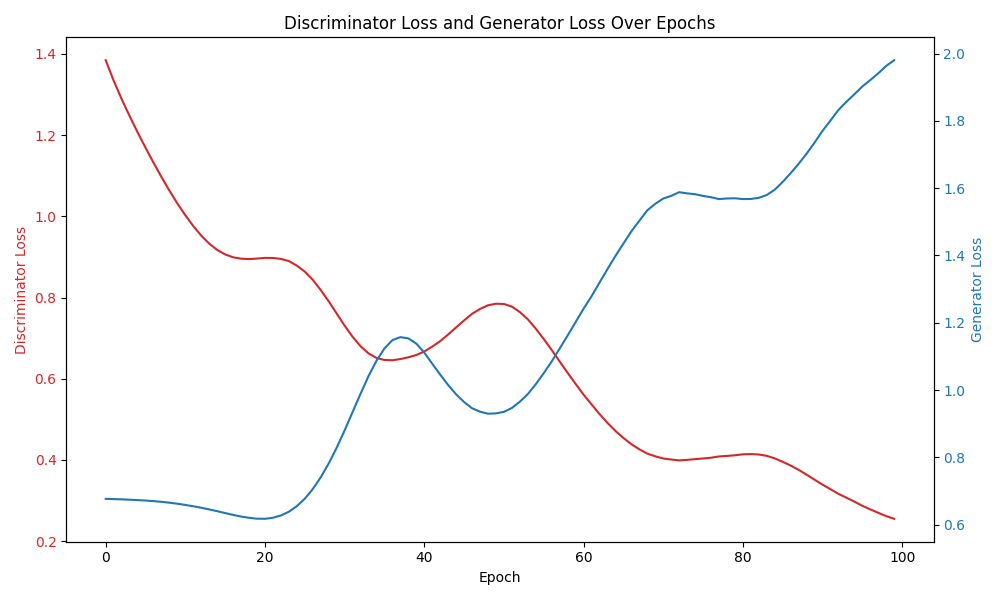}}
\\
\adjustbox{scale=0.85}{(a) Conventional GAN}
\\
\\
\adjustbox{scale=0.325}{\includegraphics{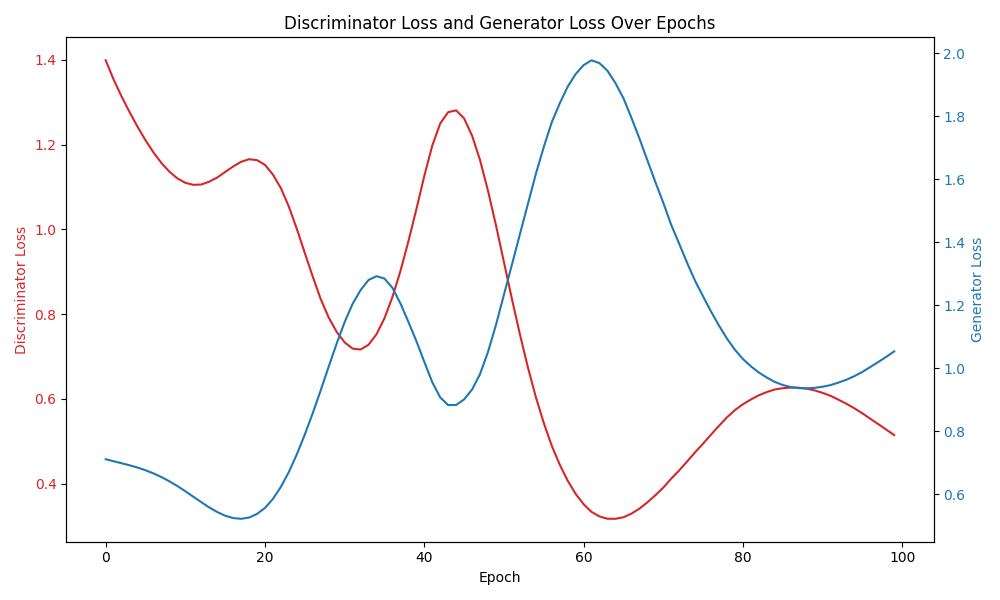}}
\\
\adjustbox{scale=0.85}{(b) Dropout-GAN}
\end{tabular}
\caption{Discriminator vs generator loss}\label{fig:DvsGloss}
\end{figure}

Training the generator against multiple discriminators proved to be effective. 
Figures~\ref{fig:DvsGloss}(a) illustrate  
the discriminator vs generator loss for a conventional GAN,
whereas Figures~\ref{fig:DvsGloss}(b) illustrate  
the discriminator vs generator loss for our Dropout-GAN architecture.
For the conventional GAN, the discriminator overwhelms 
the generator, as indicated by a significant decline in discriminator loss.
In contrast, for the Dropout-GAN architecture, the discriminator and generator losses remain comparatively  
evenly matched---we see that the generator does not get overwhelmed by the discriminator even after~100 epochs. In contrast, mode collapse begins after~25 epochs for a conventional GAN. Even when the discriminator(s) seem to have the advantage in terms of discriminator loss vs generator loss, the generator in a Dropout-GAN is able to adapt and remain roughly on par with its competition, in terms of loss. 

Note that we do not segregate humans and bots during Dropout-GAN training, similar to our conventional GAN training. Hence, we use both human and 
bot samples for training. 
But in contrast to our conventional GAN training, during Dropout-GAN training, 
the discriminators were only tested on whether they 
are able to distinguish the generated samples from the real samples, where the ``real'' samples 
could be either humans or bots. Thus, the generator was evaluated on whether it could produce 
convincing synthetic human and bot samples, and the discriminator loss was dependent on whether they could differentiate between real data versus the synthetic data generated by the generator. In the conventional GAN setup, in contrast, the emphasis was on whether the discriminator could differentiate between human users and bot users, working on a dataset containing both real data, and synthetic data with synthetic labels. 
Note that at the end of the Dropout-GAN training process, we obtain a best-performing generator G* that is resilient to mode collapse even when training against our best-performing discriminator, D*. 

Initially, we want to train a generator that can create convincing duplicates 
of both humans and bots. Once we have a functional generator, we use the best-performing 
discriminator D* from our original GAN implementation to train against this new generator, to further 
refine its capabilities and enhance its performance when detecting social media bots. 
Using this approach, we retain the early performance gains of our discriminator 
(and consequently ensure that it, in turn, does not get overwhelmed by the generator) 
as well as create a generator that can accurately model bot behavior. 
This allows the model to accurately classify users as humans or bots, 
regardless of their ratio in a test sample.

\section{Dropout-GAN Results}\label{chap:res}

In this section, we evaluate our Dropout-GAN discriminator architecture 
for bot detection, based on the MGTAB dataset. 
As mentioned above, we use an~80-10-10 training-test-validation split,
and hence our classification results are based
on~1050 users (746 humans, 304 bots on average in~10 runs) out of the~10,499 in MGTAB.

\begin{figure}[!htb]
\centering
\adjustbox{scale=0.5}{\includegraphics{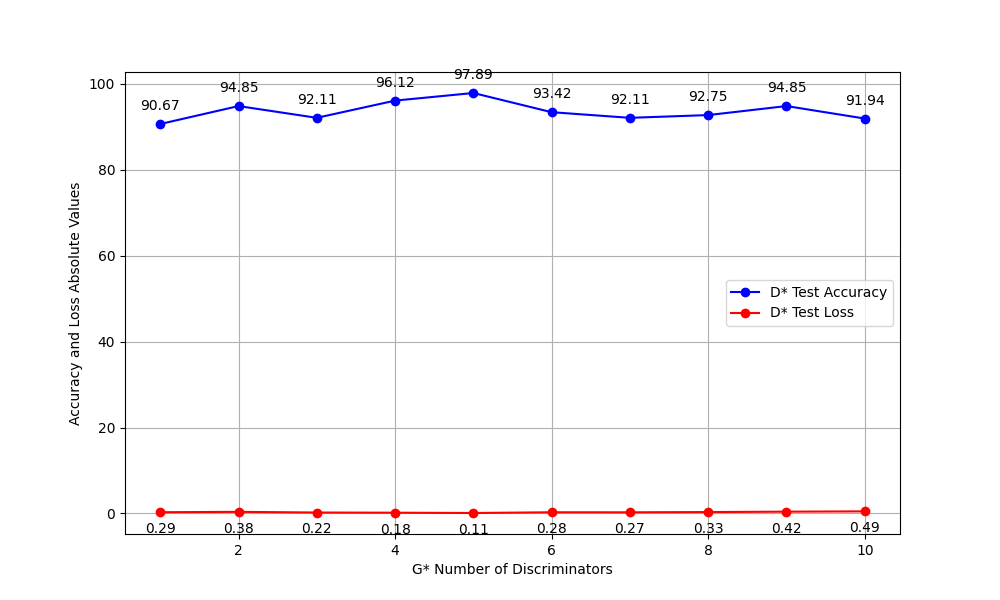}}
\caption{D* test accuracy and loss vs number of discriminators when training G*}
\label{BestComboOfDG}
\end{figure}

The results in Figure~\ref{BestComboOfDG}
show the test accuracy and loss when training the discriminator model D* 
from our conventional GAN 
against the G* obtained from a Dropout-GAN, with the number of discriminators 
in the Dropout-GAN ranging from~1 to~10.
We observe that five discriminators training against one generator
yields the best results.

We also compare the performance of our best-performing Dropout-GAN discriminator D* to 
standard feature-based learning techniques; specifically, SVM, $k$-NN, MLP, 
and Random Forest. We test these standard classifiers on the same benchmark MGTAB dataset.
The results of these classification experiments are summarized in Table~\ref{tab:performance}.
We observe that our Dropout-GAN approach outperforms all of the standard techniques tested by a wide margin.
Our Dropout-GAN approach achieves an accuracy of~99.3\%, as compared to a best result of slightly 
over~90\%\ for the standard techniques.

\begin{table}[!htb]
\centering
\caption{Comparison to selected baseline techniques}\label{tab:performance}
\adjustbox{scale=0.85}{
\begin{tabular}{c|cccc}
\midrule\midrule
Technique & Accuracy & Precision & Recall & F1-score \\ \midrule
$k$-NN & 82.94 & 79.20 & 76.88 & 77.88 \\ 
SVM & 88.62 & 86.08 & 85.27 & 85.66 \\ 
MLP & 89.22 & 86.89 & 85.90 & 86.37 \\ 
RF & 90.49 & 88.49 & 87.54 & 88.00 \\ 
Dropout-GAN & \textbf{99.3} & \textbf{99.23} & \textbf{99.10} & \textbf{99.00} \\ 
\midrule\midrule
\end{tabular}
}
\end{table}

Table~\ref{tab:performance_prevwork} summarizes the results of our best performing 
Dropout-GAN discriminator to previous work involving the same MGTAB dataset. 
Our approach outperforms all previous work, for each of the metric shown in the table. 

\begin{table}[!htb]
\centering
\caption{Comparison to previous work}\label{tab:performance_prevwork}
\adjustbox{scale=0.85}{
\begin{tabular}{l|cccc}
\midrule\midrule
Technique & Accuracy & Precision & Recall & F1-score \\
\midrule
OS-GNN~\cite{shi2023over}   & 87.18    & 85.39     & ---      & ---  \\
RF-GNN~\cite{shi2023rf}    & 87.86    & ---         & ---      & 83.99  \\
HOFA~\cite{ye2023hofa}      & 88.68    & ---         & ---      & 79.21  \\
RGT~\cite{shi2023mgtab}       & 92.12    & 88.08     & 86.64  & 90.41 \\
MSGS~\cite{shi2023muti}      & 96.59    & ---         & ---      & 96.27 \\
Dropout-GAN (this paper)  & \textbf{99.3} & \textbf{99.23} & \textbf{99.10} & \textbf{99.00} \\ 
\midrule\midrule
\end{tabular}
}
\end{table}

\begin{figure}[!htb]
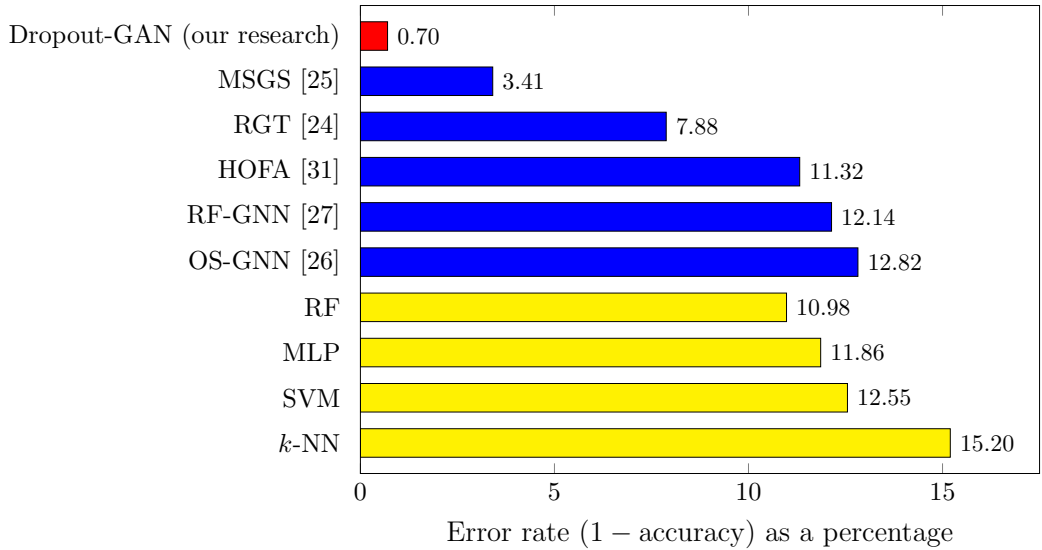

\centering
\input figures/barCompare2.tex
\caption{Error rate comparisons}\label{fig:bar}
\end{figure}

In Figure~\ref{fig:bar} we provide a bar graph that
summarizes the error rate results for our Dropout-GAN approach 
in comparison to previous research, as well as in comparison to
the baseline techniques considered
in this paper. We obtain a nearly~3\%\ improvement over the 
best previous work, and a more than~10\%\ improvement, as compared
to the best of the baseline techniques.

We also tested the best-performing discriminator D* when augmenting the
dataset using the best-performing generator G* 
from the Dropout-GAN framework. Our goal is to determine the effect
of GAN-based data augmentation. We find that varying the ratio of real data to augmented data 
does not yield large differences---varying from~50\%\ synthetic data to~100\%\ synthetic data, 
the test accuracy and test loss only changes by~0.08, based on the mean value of~10 experiments. 
Below~50\%\ synthetic data, 
the best-performing discriminator proved resilient to ``noise'' and displayed performance in line with 
the results given in Tables~\ref{tab:performance_prevwork} and~\ref{tab:performance}, above.

\begin{figure}[!htb]
\centering
\adjustbox{scale=0.4}{\includegraphics{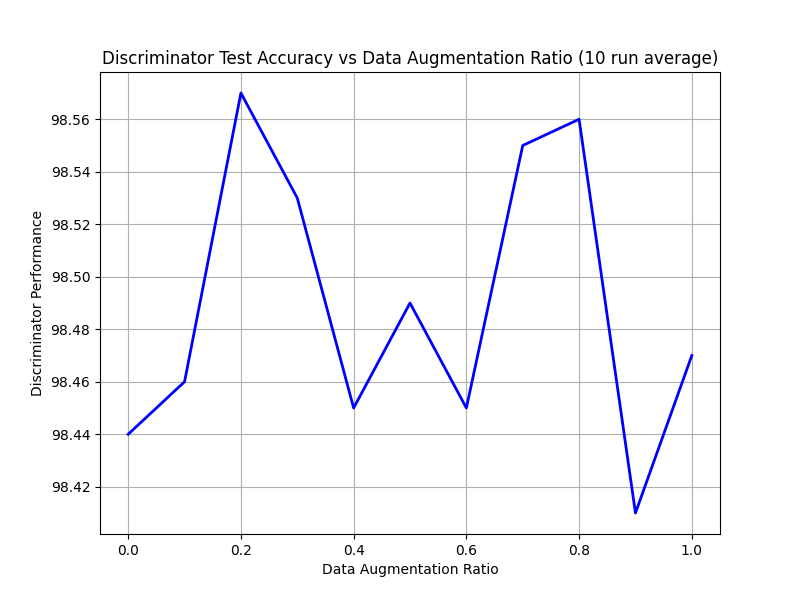}}
\caption{D* test accuracy vs data augmentation ratio}
\label{AccAugmentation}
\end{figure}

Through additional experimentation, we explored the scenario of multiple discriminators 
training against one generator in the Dropout-GAN framework.
For a sufficient number of discriminators, we find that a trained generator G* 
is able to consistently confound the best-performing discriminator D*,
when repurposing D* to distinguish between real data and G*-generated samples, and testing its accuracy in this classification.  
Specifically, in this scenario, the discriminator's accuracy plummeted to~48\%\ when five
or more discriminators are used. 
This suggests that the generator may be able to generate new kinds of bot behavior 
(and potentially, human behavior) that the discriminator cannot classify,
potentially suggesting possible behavior of social media bots in the future.  
Table~\ref{tab:DG_test_results} provides additional details on these experiments.

\begin{table}[!htb]
\centering
\caption{Best performing discriminator vs best performing generator}\label{tab:DG_test_results}
\adjustbox{scale=0.85}{
\begin{tabular}{c|cc}
\midrule\midrule
Discriminators & \multirow{2}{*}{D* test accuracy} & \multirow{2}{*}{D* test loss} \\
when training G* &  &  \\
\midrule
 \zz1 & 0.615845 & 0.770240 \\
 \zz2 &  0.981028 & 0.418309 \\
 \zz3 & 0.981028 & 0.667282 \\
 \zz4 & 0.981028 & 0.628062 \\
 \zz5 & 0.481028 & 2.116829 \\
 \zz6 & 0.481028 & 2.060582 \\
 \zz7 & 0.481028 & 1.515876 \\
 \zz8 & 0.481028 & 1.269565 \\
 \zz9 & 0.481028 & 2.246191 \\
    10 & 0.481028 & 6.534316 \\
\midrule\midrule
\end{tabular}
}
\end{table}

Ideally, a technique should be able to predict whether an account is controlled by a human or bot  
as soon as possible. We created an experiment to examine the performance of the techniques under scrutiny, by segregating accounts into percentile categories based on the date of creation and then testing the techniques on one particular category at a time. 

\begin{figure}[!htb]
\centering
\adjustbox{scale=0.5}{\includegraphics{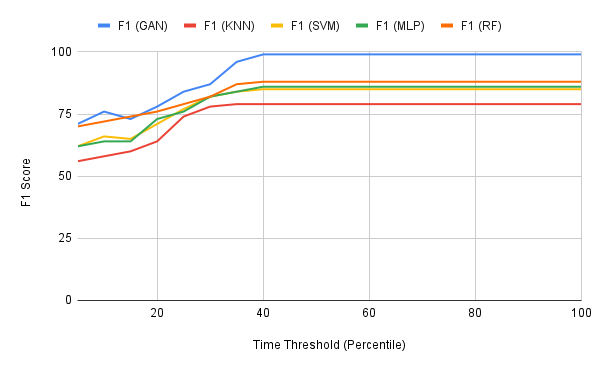}}
\caption{F1 score on $n^{\thth}$ percentile based on ``created date'' test sets}
\label{TimePerformance}
\end{figure}

Figure~\ref{TimePerformance} shows the results of this experiment. As expected, all techniques examined perform sub-optimally when tested on newer accounts since they do not have enough posts or interactions with other accounts to make a definitive decision, leading to a higher error rate in classification.  As time passes and the accounts age, the performance improves---we see a marked improvement in classification accuracy across all techniques once we test against accounts in the~$35^{\thth}$ percentile (or older) with respect to the ``created date''. Predictably, once enough time has passed, the techniques can reliably classify accounts into humans or bots, and more information does not help the classification process further, which accounts for the eventual plateauing of performance.

We also quantify the impact of each prediction that our model makes. Assuming that our model is being used in some social media platform, we want to determine the consequences of automatically banning that particular account, if classified as a bot account by our model. 
To this end, we define a new metric that we call \texttt{impact}. 
Here, \texttt{impact} is defined as the product of the number of followers, 
$F$, and the number of posts, $P$, normalized over all candidate accounts, that is,
\[
\mbox{\texttt{impact}} = \frac{F P}{\displaystyle\sum_{i=1}^{n} F_i P_i}
\]
where the sum is over all users

We can now calculate the \texttt{impact\uuu mitigation} of classifications across the testing set. If a model correctly classifies a particular account, we define the impact associated with that particular account as ``mitigated'', i.e., if that particular account was a bot, then banning that account results in a positive \texttt{impact}. In these cases, we will add the \texttt{impact} associated with the account to the \texttt{impact\uuu mitigation}. On the other hand, if we misclassify a human as a bot and ban them, then the impact is negative---we do not want to ban real users by accident, and we do not want bot accounts to remain undetected. So, in the cases of misclassification,  we subtract the \texttt{impact} associated with this account from the total impact. Thus, we have
\[
\mbox{\texttt{impact\uuu mitigation}} = \sum_{i=1}^{n} \mbox{\texttt{impact}}_i
\]

\begin{figure}[!htb]
\centering
\adjustbox{scale=0.5}{\includegraphics{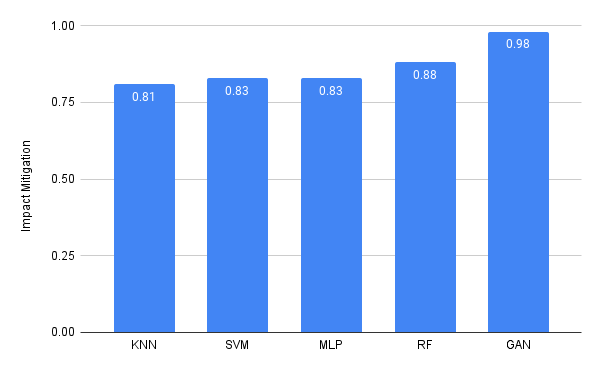}}
\caption{Impact mitigation comparison}
\label{ImpactPerformance}
\end{figure}

Note that~$-1\leq\texttt{impact\uuu mitigation}\leq 1$, since misclassifying every sample in the test set will have an \texttt{impact\uuu mitigation} of~$-1$, while correctly classify every account, then we will have an \texttt{impact\uuu mitigation} of~$1$. Figure~\ref{ImpactPerformance} illustrates the performance of various techniques with respect to this \texttt{impact\uuu mitigation} metric. Again, we see that our GAN-based approach performs best.

\begin{figure}[!htb]
\centering
\adjustbox{scale=0.5}{\includegraphics{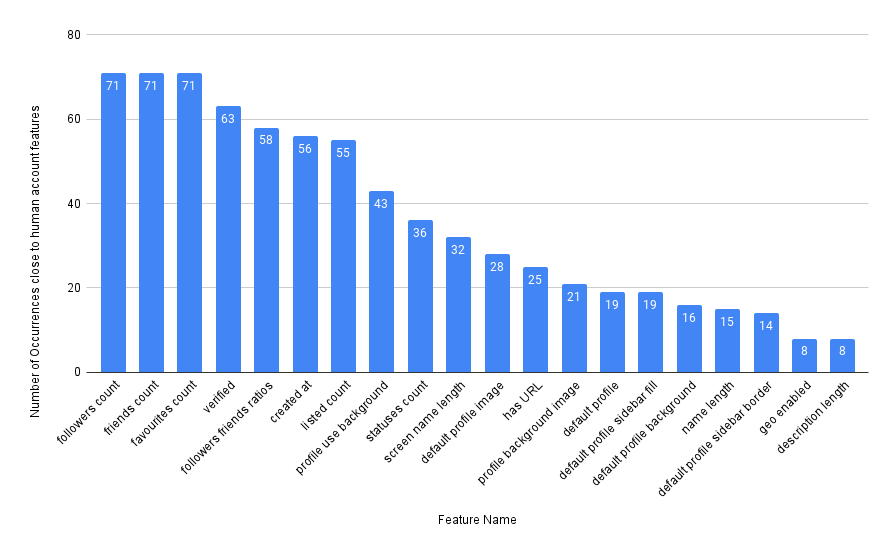}}
\caption{Generator features close to human features}
\label{closeness}
\end{figure}

We also explored the features that our best-performing generator G* is relying on to confound our best-performing discriminator D*. To this end, we examined the features that were ``close'' to real human accounts, where we define close as being within 5\%\ of the average value for a human account, for a particular feature.
The results of these experiments are summarized in Figure~\ref{closeness}.

From Figure~\ref{closeness}, we observe
that the specific attributes that contribute the most to human-like classification are 
the counts of followers, friends, and favorites. Our D* discriminator appears to rely heavily on these attributes to differentiate between human users and automated bots. Upon further investigation, we found that these three attributes were present in all instances of false positives observed during the D* testing phase. 
Additionally, an account's ``verified'' status on Twitter has historically served as a reliable indicator of its human authenticity, although this has certainty diminished with the advent of the option to purchase ``verified'' status. The follower-friend ratio and the account's age closely follow in significance as criteria aiding in the classification process.  
These results demonstrate that the generator trained in the Dropout-GAN architecture is able to produce 
synthetic bot and human behavior that is difficult to distinguish from their respective real counterparts.  

Our results indicate that the Dropout-GAN has the potential to simulate sophisticated bots 
that are difficult to detect using methods that work well to detect bots based on current bot data.
Overall, our results demonstrate that we can detect social media bots effectively, 
while there also exist mechanisms to potentially confound current state-of-the-art models.

\section{Conclusion and Future Work}\label{chap:conc}

In this paper, we proposed a novel approach to bot detection using Dropout-GANs. 
Our best performing discriminator D* outperformed state-of-the-art techniques in terms of accuracy and hence provides 
a promising approach for detecting current social media bots.

We leveraged using multiple discriminators to train one generator in the Dropout-GAN, then trained our best performing discriminator D* to detect social media bots by training it to distinguish 
between genuine user behavior and synthetic bot behavior produced by the best performing generator G*. This allows the model to 
identify patterns in the data that are characteristic of bot behavior. Perhaps most intriguing of all, we 
also demonstrated that the generator in the Dropout-GAN architecture can be used to model social media 
bots by training it to produce synthetic bot behavior that resembles real bot behavior.

Overall, our approach provides a promising direction for detecting current social media bots,
and points to directions for future research into the detection of more challenging bots. 
Our method can be applied to a wide range of social media platforms and can help to ensure 
that online discourse is more authentic and trustworthy. 

Future work could explore whether GANs can be applied to detecting social media bots across 
different social media platforms, since the dataset used for this work only included posts 
from one platform, i.e., Twitter. Additional future investigations may also consider the extension of our approach 
to datasets beyond social media bots to validate its applicability across different domains. Another area for exploration could be experimenting with 
more types of layers for the GAN model, such as LSTM layers or convolutional layers. 
One could also investigate how bots can reverse-engineer such models and 
overcome them with new behavior that may not be captured by our generator. 
One could also repeat the experiments with graph-based techniques such as HGT, GAT, GCN,
and so on.

\bibliographystyle{plain}

\bibliography{references.bib}

\end{document}

%% file: preamble.tex
\usepackage{amsmath,amsthm, amsfonts, amssymb, amsxtra, amsopn}
\usepackage{pgfplots}
\pgfplotsset{compat=1.13}
\usepackage{pgfplotstable}
\usepackage{graphicx,grffile}
\usepackage{multirow}
\usepackage{booktabs}
\usepackage{listings}
\usepackage{cmap}
\usepackage{colortbl}
\usepackage{adjustbox}
\usepackage{epsfig}

\usepackage[tableposition=top,font=small,skip=5pt]{caption}
\usepackage{subcaption}
\usepackage{makecell} 

\PassOptionsToPackage{hyphens}{url}

\usepackage{hyperref}
\hypersetup{colorlinks=true,linkcolor=black,citecolor=black,urlcolor=blue,filecolor=black}
\hypersetup{pdfpagemode=UseNone,pdfstartview=}

\usepackage{enumitem}
\setlist[itemize]{noitemsep, topsep=0pt}

\advance\oddsidemargin by -0.35in
\advance\textwidth by 0.7in

\advance\topmargin by -0.4in
\advance\textheight by 0.8in

\long\def\symbolfootnotetext[#1]#2{\begingroup%
\def\thefootnote{\fnsymbol{footnote}}\footnotetext[#1]{#2}\endgroup}

\newcommand\dunderline[3][-1pt]{{%
  \sbox0{#3}%
  \ooalign{\copy0\cr\rule[\dimexpr#1-#2\relax]{\wd0}{#2}}}}
\def\uuu{\kern-1pt\dunderline{0.75pt}{\phantom{M}}}

\DeclareMathOperator{\thth}{th}

\let\vvv=\v
\def\v{{\tt v}}

\def\zz{\phantom{0}}

%% file: figures/barCompare2.tex
\begin{tikzpicture}[scale=0.9, every node/.style={scale=1.0}]
\begin{axis}[ 
xbar,
width  = 0.75*\textwidth,
height = 8.5cm,
xmin=0,xmax=17.5,
xtick={0,5,10,15},
xlabel = {Error rate ($1 - \mbox{accuracy}$) as a percentage},
major y tick style = transparent,
bar shift=0pt,
bar width=12.0pt,
x tick label style={
	font=\small,
    	/pgf/number format/.cd,
   	fixed,
   	fixed zerofill,
    	precision=0},
y tick label style={
		font=\small,
		},
symbolic y coords={%
	k-NN,
	SVM,
	MLP,
	RF,
	OS-GNN,
	RF-GNN,
	HOFA,
	RGT,
	MSGS,
	Ours
	},
yticklabels={%
	$k$-NN,
	SVM,
	MLP,
	RF,
	OS-GNN~\cite{shi2023over},
	RF-GNN~\cite{shi2023rf},
	HOFA~\cite{ye2023hofa},
	RGT~\cite{shi2023mgtab},
	MSGS~\cite{shi2023muti},
	Dropout-GAN (our research),
	},
ytick=data,
enlarge y limits=0.075,
nodes near coords={\pgfmathfloatifflags{\pgfplotspointmeta}{0}{}{\pgfmathprintnumber{\pgfplotspointmeta}}},
every node near coord/.append style={
	font=\footnotesize,
	/pgf/number format/.cd,
	fixed,
	fixed zerofill,
	precision=2},
ytick=data,
]
\addplot [fill=yellow,opacity=1.00]
coordinates {
(15.20,k-NN)
(0.0,SVM)
(0.0,MLP)
(0.0,RF)
(0.0,OS-GNN)
(0.0,RF-GNN)
(0.0,HOFA)
(0.0,RGT)
(0.0,MSGS)
(0.0,Ours)
};
\addplot [fill=yellow,opacity=1.00]
coordinates {
(0.0,k-NN)
(12.55,SVM)
(0.0,MLP)
(0.0,RF)
(0.0,OS-GNN)
(0.0,RF-GNN)
(0.0,HOFA)
(0.0,RGT)
(0.0,MSGS)
(0.0,Ours)
};
\addplot [fill=yellow,opacity=1.00]
coordinates {
(0.0,k-NN)
(0.0,SVM)
(11.86,MLP)
(0.0,RF)
(0.0,OS-GNN)
(0.0,RF-GNN)
(0.0,HOFA)
(0.0,RGT)
(0.0,MSGS)
(0.0,Ours)
};
\addplot [fill=yellow,opacity=1.00]
coordinates {
(0.0,k-NN)
(0.0,SVM)
(0.0,MLP)
(10.98,RF)
(0.0,OS-GNN)
(0.0,RF-GNN)
(0.0,HOFA)
(0.0,RGT)
(0.0,MSGS)
(0.0,Ours)
};
\addplot [fill=blue,opacity=1.00]
coordinates {
(0.0,k-NN)
(0.0,SVM)
(0.0,MLP)
(0.0,RF)
(12.82,OS-GNN)
(0.0,RF-GNN)
(0.0,HOFA)
(0.0,RGT)
(0.0,MSGS)
(0.0,Ours)
};
\addplot [fill=blue,opacity=1.00]
coordinates {
(0.0,k-NN)
(0.0,SVM)
(0.0,MLP)
(0.0,RF)
(0.0,OS-GNN)
(12.14,RF-GNN)
(0.0,HOFA)
(0.0,RGT)
(0.0,MSGS)
(0.0,Ours)
};
\addplot [fill=blue,opacity=1.00]
coordinates {
(0.0,k-NN)
(0.0,SVM)
(0.0,MLP)
(0.0,RF)
(0.0,OS-GNN)
(0.0,RF-GNN)
(11.32,HOFA)
(0.0,RGT)
(0.0,MSGS)
(0.0,Ours)
};
\addplot [fill=blue,opacity=1.00]
coordinates {
(0.0,k-NN)
(0.0,SVM)
(0.0,MLP)
(0.0,RF)
(0.0,OS-GNN)
(0.0,RF-GNN)
(0.0,HOFA)
(7.88,RGT)
(0.0,MSGS)
(0.0,Ours)
};
\addplot [fill=blue,opacity=1.00]
coordinates {
(0.0,k-NN)
(0.0,SVM)
(0.0,MLP)
(0.0,RF)
(0.0,OS-GNN)
(0.0,RF-GNN)
(0.0,HOFA)
(0.0,RGT)
(3.41,MSGS)
(0.0,Ours)
};
\addplot [fill=red,opacity=1.00]
coordinates {
(0.0,k-NN)
(0.0,SVM)
(0.0,MLP)
(0.0,RF)
(0.0,OS-GNN)
(0.0,RF-GNN)
(0.0,HOFA)
(0.0,RGT)
(0.0,MSGS)
(0.7,Ours)
};
\end{axis}
\end{tikzpicture}